\documentclass[letterpaper]{article} 
\usepackage{aaai25}  
\usepackage{times}  
\usepackage{helvet}  
\usepackage{courier}  
\usepackage[hyphens]{url}  
\usepackage{graphicx} 
\usepackage{natbib}  
\usepackage{caption} 
\usepackage[ruled]{algorithm2e}
\usepackage{booktabs}  
\usepackage{amssymb}
\usepackage{newfloat}
\usepackage{listings}
\usepackage{microtype}
\usepackage{subfig}
\usepackage{inconsolata}

\usepackage{multirow}
\usepackage{bm}
\usepackage{amsmath,epsfig}
\usepackage{lipsum}
\usepackage{floatrow}
\usepackage{tcolorbox} 

\usepackage[switch]{lineno}

\title{EPERM: An Evidence Path Enhanced Reasoning Model\\for Knowledge Graph Question and Answering}

\author{Xiao Long$^{1}$, Liansheng Zhuang$^{1}$\thanks{Corresponding author.}, Aodi Li$^{1}$, Minghong Yao$^{1}$, Shafei Wang$^{2}$}

\affiliations{
$^{1}$University of Science and Technology of China, Hefei 230026, China \\$^{2}$Peng Cheng Laboratory, Shenzhen 518000, China \\ longxiao1997@mail.ustc.edu.cn; lszhuang@ustc.edu.cn
}

\begin{document}

\maketitle

\begin{abstract}
Due to the remarkable reasoning ability, Large language models (LLMs) have demonstrated impressive performance in knowledge graph question answering (KGQA) tasks, which find answers to natural language questions over knowledge graphs (KGs). To alleviate the hallucinations and lack of knowledge issues of LLMs, existing methods often retrieve the question-related information from KGs to enrich the input context. However, most methods focus on retrieving the relevant information while ignoring the importance of different types of knowledge in reasoning, which degrades their performance. To this end, this paper reformulates the KGQA problem as a graphical model and proposes a three-stage framework named the Evidence Path Enhanced Reasoning Model (EPERM) for KGQA. In the first stage, EPERM uses the fine-tuned LLM to retrieve a subgraph related to the question from the original knowledge graph. 
In the second stage, EPERM filters out the evidence paths that faithfully support the reasoning of the questions, and score their importance in reasoning.
Finally, EPERM uses the weighted evidence paths to reason the final answer. Since considering the importance of different structural information in KGs for reasoning, EPERM can improve the reasoning ability of LLMs in KGQA tasks.  
Extensive experiments on benchmark datasets demonstrate that EPERM achieves superior performances in KGQA tasks.
\end{abstract}

\section{Introduction}

Question answering over knowledge graph (KGQA) has garnered significant attention in recent years. It aims to find answers for natural language questions based on knowledge graphs (KGs), such as Freebase~\cite{bollacker2008freebase} and Wikidata~\cite{vrandevcic2014wikidata}, which are built from a large number of triplets consisting of (head entity, relation, tail entity). Since the natural language questions often contain compositional semantics~\cite{lan2022complex}, exactly understanding the semantic information in the question and identifying the structured knowledge in KGs is very important for KGQA tasks.

Recently, as large language models (LLMs)~\cite{openai2023gpt, hadi2023survey} have demonstrated impressive ability to understand natural language and reasoning abilities in many NLP tasks, LLMs have also shown impressive performance in knowledge graph question answering tasks~\cite{jiang2022unikgqa}. Currently, retrieval-augmented methods~\cite{wu2023retrieve, sun2023think, ding2024enhancing} are popular ones that combine LLMs and KGs for KGQA tasks. Usually, they involve two steps. First, they retrieve the question-related triplets or paths as contextual knowledge from the raw KGs. Then, they leverage these contexts for the LLM to generate the answers. Although retrieval-augmented methods exploit the ability of LLMs for reasoning and have achieved promising results in KGQA tasks~\cite{wu2023retrieve, sun2023think}, they still suffer from the following issues. First, they usually treat the different retrieval information equally to reason the answer, ignoring the differences between retrieved information. Second, the retrieval-augmented generation methods usually treat the retrieval and reasoning processes separately in model learning. The coupling between the retrieval and reasoning processes of the model is low, and there is a lack of a unified framework to model KGQA tasks.

Inspired by the above insights, this paper reformulates the knowledge graph question answering task as a probabilistic graphical model~\cite{koller2009probabilistic}, and proposes a novel framework named Evidence Path Enhanced Reasoning Model (EPERM), which considers the importance of different structural information when reasoning the question answers. Our EPERM involves the subgraph retrieval stage, the evidence path finding stage, and the answer prediction stage. Specifically, in the first stage, the subgraph retriever is proposed to retrieve a subgraph related to the question from the original knowledge graphs. In the second stage, the proposed evidence path finder first generates a series of weighted plans that faithfully support the reasoning of the questions. Then it scores and filters out the weighted evidence path in the subgraph based on the weighted plans. In the final stage, the answer predictor is proposed to use the weighted evidence path to reason the final answer. Since the weight of each evidence path represents the importance of the structural information for reasoning the problem, EPERM can better leverage them to reason the answer. In addition, we design joint fine-tuning strategies to learn the parameters in the retrieval and reasoning processes. Finally, since the entire question-answering process is described as a probabilistic graphical model, EPERM exhibits greater coupling between the retrieval and reasoning stages. Our contributions can be summarized as follows. 
\begin{itemize} 
    \item This paper reformulates the Knowledge Graph Question Answering (KGQA) problem as a graphical model and proposes a novel framework named the Evidence Path Enhanced Reasoning Model (EPERM), which leverages the reasoning capabilities of Large Language Models (LLMs) and the structure information in KGs. By considering the varying importance of the structural information, our EPERM can achieve better reasoning abilities for KGQA tasks. 
    \item A joint fine-tuning strategy is proposed to improve the reasoning abilities of LLMs guided by the graphical model for KGQA. Compared with previous works on KQGA tasks, EPERM is more unified and exhibits greater coupling between the retrieval and reasoning stages.
    \item Extensive experiments on two benchmark datasets demonstrate that EPERM achieves superior performances in KGQA tasks and significantly outperforms all types of state-of-the-art methods on these two datasets. Especially on WebQSP, EPERM achieves a 3.6 \% relative improvement in Hit@1 score compared to the state-of-the-art methods.
\end{itemize}

\section{Related Work}
\textbf{Knowledge Graph Question Answering.}
Conventional KBQA solutions can be categorized into three types: Semantic Parsing-based (SP-based) methods, Information Retrieval-based (IR-based) methods, and Embedding-based methods. SP-based methods parse the question into a structural query (e.g., SPARQL) which can be executed by a query engine to get answers~\cite{lan2022complex}. ArcaneQA~\cite{gu2022arcaneqa} dynamically generates the query based on results from previous steps. RnG-KBQA~\cite{ye2021rng} first enumerate all possible queries and then rank them to get the final output. These methods heavily rely on the quality of generated queries. If the query is not executable, no answers will be generated. DECAF~\cite{donahue2014decaf} combines semantic parsing and LLMs reasoning to jointly generate answers, which also reach salient performance on KGQA tasks. However, these methods need to annotate expensive logic forms as supervision or are limited to narrow domains with a few logical predicates~\cite{lan2022complex}. KG embedding, which aims to encode entities and relations into a continuous vector space~\cite{bordes2013translating,long2022neural, sun2019rotate, long2024fact}, and its effectiveness has been validated in knowledge graph question answering (KGQA) tasks. Embedding-based methods model the entities and relations in embedding space and design special model architectures to reason answers. KV-Mem~\cite{miller2016key} adopts a Key-Value memory network to store triples for reasoning. EmbedKGQA~\cite{saxena2020improving} and NSM~\cite{he2021improving} utilize the sequential model to mimic the multi-hop reasoning process. IR-based methods primarily retrieve relevant factual triples or text from Knowledge Graphs (KGs) based on natural language questions and then design special model architectures to reason answers. Early works adopt the page rank or random walk algorithm to retrieve subgraphs from KGs for reasoning~\cite{sun2018open}. Recently, to integrate LLMs for KGQA, retrieval augmented methods~\cite{jiang2022unikgqa, luo2023reasoning} aim to leverage the LLMs to reason on the retrieved facts from the KGs to improve the reasoning performance. For example, UniKGQA~\cite{jiang2022unikgqa} unifies the graph retrieval and reasoning process into a single model with LLMs. ToG~\cite{sun2023think} uses LLM as an agent to iteratively perform beam search on knowledge graphs to find answers. RoG~\cite{luo2023reasoning} uses LLM to generate relation plans, which are used to retrieve the relative facts from raw KGs for LLMs to conduct faithful reasoning. However, these methods treat the different retrieval information equally to reason the answer, ignoring the differences between retrieved information. EPERM proposes to retrieve and score the evidence paths, which consider the different importance of the structural information for better reasoning the answers. 

\noindent \textbf{Large Language Models.} 
With the launch of ChatGPT and GPT-4~\cite{openai2023gpt}, displaying the prowess of decoder-only large language models (LLMs) with a vast number of parameters that exhibit emergent phenomena, many traditional NLP tasks are becoming simplified~\cite{hadi2023survey}. Subsequently, open-source models like Llama-2-7B~\cite{touvron2023llama}, ChatGLM2-6B~\cite{zeng2022glm} and Qwen-Chat~\cite{bai2023qwen} emerged and can be supervised fine-tuned (SFT) using instruction-tuning technologies~\cite{zhang2023fc} such as LoRA~\cite{hu2021lora}, QLoRA~\cite{dettmers2024qlora}, P-Tuning v2~\cite{liu2021p}, and Freeze~\cite{geva2020transformer}, enhancing the capabilities of LLMs for specific tasks. Additionally, Chain-of-Thought (CoT)~\cite{wei2022chain} has been shown to be effective in enhancing LLM reasoning. It creates a series of prompt instances according to reasoning logic under a few-shot learning paradigm in order to improve LLM’s performance on complex tasks. In this paper, EPERM employs the instruction-tuning technique to fine-tune open-source LLMs, which consists of the subgraph retriever, evidence path finder, and answer predictor. All the modules in EPERM are joint fine-tuning to learn the parameters.

\section{Methodology}
Overall, the framework of the Evidence Path Enhanced Reasoning Model (EPERM) is shown in Figure~\ref{p2}, which contains the subgraph retriever module, the evidence path finder module and the answer predictor module. In the first stage, EPERM first uses the fine-tuned LLM to retrieve a subgraph related to the question from the original knowledge graph. In the second stage, the proposed evidence path finder first generates a series of weighted plans that faithfully support the reasoning of the questions. Then it scores and filters out the weighted evidence path in the subgraph based on the weighted plans. Finally, EPERM uses the evidence paths with their importance score to reason the final answer. In the following subsections, we first formally define the KGQA task. Then, we introduce the details of the proposed method.

\subsection{Problem Definition}
A knowledge graph typically consists of a set of triples, denoted by $\mathcal{G}=\{(e, r, e^{'})|e, e^{'} \in E, r \in R\}$, where $E$ and $R$ denote the entity set and relation set, respectively. Knowledge Graph Question Answering (KGQA) is a typical reasoning task based on KGs. Given a natural language question $Q_{n}$ and a KG $\mathcal{G}$, the task aims to design a function $f$ to predict answers $\mathcal{A}_n$ based on knowledge graph $\mathcal{G}$, i.e., $\mathcal{A}_n = f(Q_n,\mathcal{G})$. Following previous works~\cite{zhang2022subgraph} , we assume the topic entities $\mathcal{T}_n$ mentioned in $Q_n$ and answers $\mathcal{A}_n$ are labeled and linked to the corresponding entities in $\mathcal{G}$, i.e., $\mathcal{T}_{n}, \mathcal{A}_{n} \subseteq E$. Additionally, given a question $Q_n$ and an answer $A_n$, the i-th evidence path instance connecting $e_{Q_n}$ and $e_{A_n}$ in KGs is defined as $P_{\mathcal{P}_{i}(e_{Q_n}, e_{A_n})} = e_{Q_n} \xrightarrow{r^{i}_1} e_1 \xrightarrow{r^{i}_2} \cdots \xrightarrow{r^{i}_l} e_{A_n}$. The corresponding i-th plan $p^{i} = \{r^{i}_1, r^{i}_2, \cdots r^{i}_l\}$ can be considered a faithful plan for reasoning the question $Q_n$. And $\mathcal{P}_{i}(e_{Q_n}, e_{A_n})=\{p^i | i = 1,\cdots, s\}$ is a set of plans to the question $Q_n$, which $s$ is the number of plans.

\begin{figure}[htbp] 
\centering
{\includegraphics[height=0.4\textwidth]{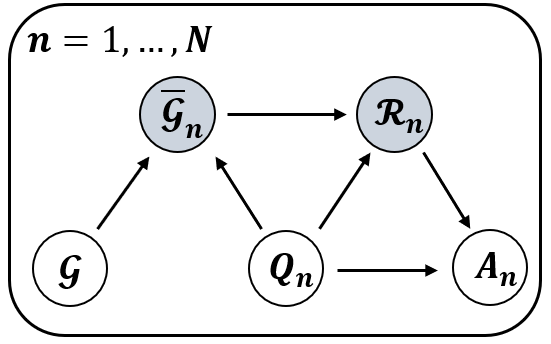}}
\caption{The directed graphical model of KGQA tasks.}
\label{p1} 
\end{figure}

\subsection{The EPERM Framework}
We start by formalizing the knowledge graph question answering in a probabilistic way, Given a question $Q_n$ and its answers $\mathcal{A}_n$, we formalize the KBQA problem as to model the probabilistic distribution $P(\mathcal{A}_n|\mathcal{G}, Q_n)$. We introduce two latent variables: a question-related subgraph $\overline{\mathcal{G}}$ and a series of evidence paths $\mathcal{R}_n$ help to reason the question $Q_n$. Then the KGQA task can be reformulated as a probabilistic graphical model in Figure~\ref{p1}. Based on the independence among the variables in the directed graph and following the d-separation principle~\cite{pearl2009causality}, the proposed model (objective distribution) $P(\mathcal{A}_n|\mathcal{G}, Q_n)$ can be reformulated as below (we include these details in the Appendix):
\begin{align}
\label{eq1}
    P_{\theta}(\mathcal{A}_n|\mathcal{G}, Q_n) = \Sigma_{\mathcal{R}_n} \Sigma_{\overline{\mathcal{G}}} P_{\theta}(\mathcal{A}_n|\mathcal{R}_n, Q_n) \nonumber \\ P_{\theta}(\mathcal{R}_n|\overline{\mathcal{G}}, Q_n) P_{\theta}(\overline{\mathcal{G}}|\mathcal{G}, Q_n).
\end{align}
In the above equation, the proposed EPERM can be divided into three parts. The first part is the subgraph retriever module, which is described by $P_{\theta}(\overline{\mathcal{G}}|\mathcal{G}, Q_n)$. It aims to retrieve a question-related subgraph from the KGs. The second part is the evidence path finder module, which is described by $P_{\theta}(\mathcal{R}_n|\overline{\mathcal{G}}, Q_n).$ Its aim is to find and evaluate evidence paths for the subsequent reasoning. Finally, the answer predictor module is formulated by $P_{\theta}(\mathcal{A}_n|\mathcal{R}_n, Q_n)$, which leverages the weighted evidence paths to reason the final answer. The following sections will provide a detailed introduction to the three modules of EPERM and its training objectives.

\begin{figure*}[htbp]
\centering
{\includegraphics[height=0.33\textwidth]{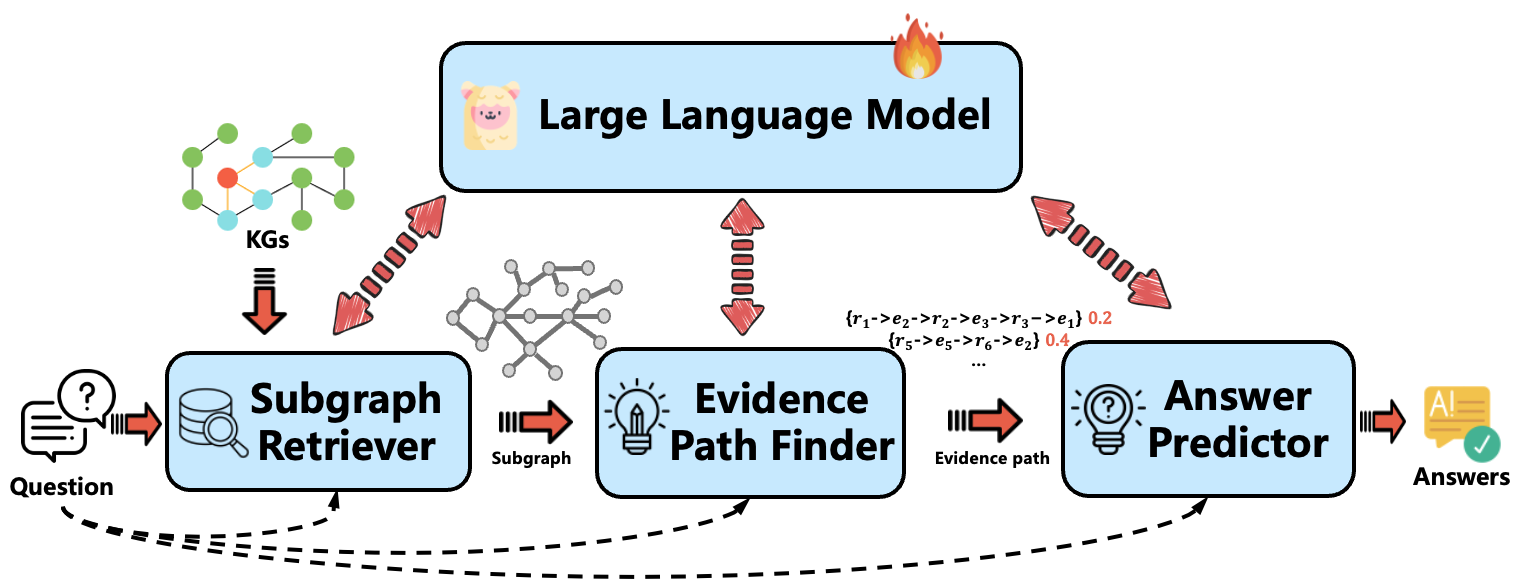}} 
\caption{Overview of the proposed EPERM framework. The subgraph retriever module aims to retrieve the question-related subgraph. The evidence path finder module aims to find and score the importance of evidence reasoning paths. The answer predictor module aims to reason the final answer based on the weighted evidence paths.}
\label{p2}
\end{figure*} 

\noindent \textbf{Subgraph retriever module.}\
The subgraph retriever module aims to calculate $P_{\theta}(\overline{\mathcal{G}}|\mathcal{G}, Q_n)$ for any $\overline{\mathcal{G}}$, which is intractable as the latent variable $\overline{\mathcal{G}}$ is combinatorial in nature. To avoid enumerating $\overline{\mathcal{G}}$, we propose to expand top-$K$ paths relevant to $Q_n$ from the topic entities. Specifically, path expansion starts from a topic entity $\mathcal{T}_n$ and follows a sequential decision process. At the beginning of the iteration, the relation expansion phase first searches out all relations $\{r^{0}_i\}^{N}_{i=1}$ linked to the topic entity $\mathcal{T}_n$. Then, we select the top $K$ relations $\{r^{0}_i\}^{K}_{i=1}$ by using the fine-tuned LLM to score the relevance of each $\{r^{0}_i\}^{N}_{i=0}$ to the question $Q_n$. 
\begin{align}
\label{eq2}
    S(Q_n, r) = LLM_{\theta}(r, Q_n).
\end{align}
The scoring procedure is completed by executing a pre-defined formal query shown in the Appendix. Then, we retrieve the corresponding tail entities $\{E_j\}^{K}_{j=1}$ connected to corresponding $K$ relations.
Normally, at $D$-th iteration, we still perform a top-$K$ beam search from current entities to get the $K$ relations. In this way, we can get the subgraph $\overline{\mathcal{G}}$ related to question $Q_n$.

\begin{algorithm}[t]
 \caption{Inference Stage}
 \label{alg:algorithm1}
 \textbf{Input}: KG $\mathcal{G}$, question $Q_n$, topic entities $\mathcal{T}_n$; \\
 \textbf{Output}: Answer $\mathcal{A}_n$; \\
 $\overline{\mathcal{G}} \leftarrow SubgraphRetriever(\mathcal{G}, Q_n)$ \\
 Evidence path \ $\mathcal{R}_{n} \leftarrow [ ]$  \\
 $\mathcal{P}_n=\{ p^1, \cdots, p^s \} \leftarrow Generator(Q_n)$ \\
 \ForEach{$e_T \in \mathcal{T}_n$} {
    $E^{e_T}_{0} \leftarrow [e_T]$ \\
    \ForEach{$p^{j} \in \mathcal{P}_n$}{
        \For{$i \leftarrow 1 \ \mathrm{to} \ \mathrm{length}(p^j) + 1$}
        {
            $E^{e_T}_{i} \leftarrow SearchAdj(e_T, r^{j}_{i-1}, \overline{\mathcal{G}})$ \\
            $E^{S}_{i} \leftarrow S(Q_n,\ E^{e_T}_{i})$ \\
            $E^{filter}_{i} \leftarrow topK(E^{S}_{i})$ \\
            $\mathcal{R}_{n}.\text{append}([E^{e_T}_{i-1}, p^{j}_{i}, E^{filter}_{i}])$ \\
            $E^{e_T}_{i} \leftarrow E^{filter}_{i}$
        }
    }
 } 
$\mathcal{A}_n \leftarrow AnswerPredictor(\mathcal{R}_n)$ 
\end{algorithm}


\noindent\textbf{Evidence path finder module.}\ This module aims to score and filter out a series of weighted evidence paths that faithfully support the reasoning of the questions. First, it generates a series of latent weighted plans $\mathcal{P}_n=\{ p^1, \cdots, p^s \}$ for answering the question $Q_n$ based on the subgraph $\overline{\mathcal{G}}$ in the previous stage. This process is defined by the distribution $P_{\theta}(\mathcal{P}_n|\overline{\mathcal{G}}, Q_n)$, which specifically modeled by a fine-tuned LLM called $Generator_{\theta}$. Specifically, the $Generator_{\theta}$ generates compositional plans by only considering the question $Q_n$ and the subgraph $\overline{\mathcal{G}}$, which allows these plans to generalize across entities in KGs. The methods for fine-tuning the generator will be explained in detail in the following sections.
To utilize the instruction-following ability of LLMs~\cite{zhang2023fc}, we design a simple instruction template that prompts LLMs to generate the compositional plan in $\mathcal{P}_n=\{ p^1, \cdots, p^s \}$ and their scores. For each compositional plan $p^j$ in the $\mathcal{P}_n$, it can be viewed as a sequence of relations $\{r^{j}_1, r^{j}_2 \cdots, r^{j}_l\}$. The $\{r^{j}_i\}^{l}_{i=1}$ is the body of the plan,
where $l$ is the max hops for the plan, and $s$ is the total number of the weighted plans for one question $Q_n$. After obtaining the weighted plans $\mathcal{P}_n$, we need to further retrieve and score the importance of evidence reasoning paths $\mathcal{R}_n$. For each plan $p^j=\{r^{j}_i\}^{l}_{i=1}$, a path tree can be induced by filling in the intermediate entities along the plan, i.e., $T(j) = (\mathcal{T}_n, r^{j}_1, E_1, \cdots, r^{j}_l, E_l)$. In general, given a head entity, the multi-semantics of the relations~\cite{long2024kgdm} often lead to multiple tail entities. Therefore, each $E_i \in \{ E_1, \cdots, E_l\}$ can be represented  $E_{i} = \{e^{m}_i\}^{V}_{m=1}$, which is an entity set and it leads to a situation where a plan connects multiple paths. However, not all entities in the path help answer the question, and we further score the relevance between question $Q_n$ and the entities by using the information surrounding them. The prompt of scoring entities shown in the Appendix.
\begin{align}
E^{S}_{i} = S(Q_n, SearchAdj(e^{m}_{i-1}, r^{j}_{i-1}, \overline{\mathcal{G}})).
\end{align}
The $SearchAdj(e^{m}_{i-1}, r^{j}_{i-1}, \overline{\mathcal{G}})$ is used to get all the adjacent entities of $e^{m}_{i-1}$ given the pre-relation $r^{j}_{i-1}$ in $\overline{\mathcal{G}}$. After obtaining the scores for all adjacent entities, we filter the top $S$ score entities to form corresponding top $S$ paths in every hop. Finally, by multiplying the scores on the paths, we get the evidence path $\mathcal{R}_{n}$.

\noindent\textbf{Answer predictor module.}\
The answer predictor module takes the question $Q_n$ and the evidence paths $\mathcal{R}_{n}$ to generate answers $\mathcal{A}_n$, which defined by $P_{\theta}(\mathcal{A}_n|\mathcal{R}_n, Q_n)$. Similarly, we design a reasoning instruction prompt to guide the fine-tuned LLMs to conduct reasoning based on the evidence path $\mathcal{R}_{n}$. The details of the prompts can be found in the Appendix. Finally, given the input of the question $Q_n$ and a raw knowledge graph $\mathcal{G}$, the pseudocode for the complete inference process is presented in Algorithm~\ref{alg:algorithm1}.

\subsection{Optimization Framework}
Next, we introduce how to optimize the EPERM framework. Since the LLMs have zero knowledge of the relations contained in KGs. Therefore, LLMs cannot directly generate weighted plans $\mathcal{P}_n$ and the evidence paths $\mathcal{R}_n$ grounded by KGs. Moreover, LLMs might not understand the evidence paths correctly and conduct reasoning based on them. To address these issues, we design a joint instruction tuning task. The objective function in equation~\ref{eq1} can be optimized by maximizing the evidence lower bound (ELBO)~\cite{hoffman2016elbo}, which is formulated as:
\begin{equation}
\label{eq5}
    \begin{aligned}
    &logP(\mathcal{A}_n|\mathcal{G}, Q_n) \ge \mathbb{E}_{\mathcal{R}_n \sim q_1}[logP(\mathcal{A}_n | \mathcal{R}_n) \\ & - D_{KL}(q_{2}(\overline{\mathcal{G}}|\mathcal{G}, Q_n) || P(\overline{\mathcal{G}}|\mathcal{G}, Q_n))] \\ &- \mathbb{E}_{\overline{\mathcal{G}} \sim q_2}[D_{KL}(q_{1}(\mathcal{R}_n|A_n) || P(\mathcal{R}_n|\overline{\mathcal{G}}, Q_n))]. 
    \end{aligned}
\end{equation}
where $q_{1}(\mathcal{R}_n|A_n)$ denotes the posterior distribution of faithful evidence paths grounded by KGs and
$q_{2}(\overline{\mathcal{G}} | \mathcal{G}, Q_n)$ denotes the posterior distribution of subgraph. Since we define how to retrieve the subgraph, the posterior distribution of the subgraph is known and the parameters in equation~\ref{eq2} can be learned in the evidence path finding stage. So, we need to optimize the first and the third items in equation~\ref{eq5}, which represent to evidence path finder module and the answer predictor module, respectively. We will provide a detailed introduction to these two parts.

\noindent\textbf{Evidence path finder module optimization.} To optimize the evidence path finder module, we aim to distill the knowledge from KGs into LLMs to generate faithful evidence paths. This can be achieved by minimizing the KL divergence with the posterior distribution of faithful evidence paths $q_{1}(\mathcal{R}_n)$, which can be approximated by the valid plans $\mathcal{P}_n$ in KGs. Specifically, given a question $Q_n$ and plans $\mathcal{P}_n$, we may retrieve many candidate answer entities $f_{Q_n}(\mathcal{P}_n)$ in the knowledge graph $\overline{\mathcal{G}}$. We select the plan, in which the ratio of the answer entity to all candidate entities is greater than the threshold value. The posterior
distribution distribution $q_{1}(\mathcal{R}_n)$ can be formally approximated as:
\begin{equation}
     q_{1}(\mathcal{R}_n) \simeq q_{1}(\mathcal{R}_n | A_n, Q_n) = 
    \begin{cases} 
    1,& \frac{N(A_n\in f_{Q_n}(\mathcal{P}_n))}{||f_{Q_n}(\mathcal{P}_n)||} \ge t \\ 
    0,& else.
    \end{cases}
\end{equation}
Therefore, the KL divergence can be calculated as
\begin{equation}
    \mathcal{L}_{find} = -\sum_{\mathcal{R}_n \in q_1(\mathcal{R}_n)} logP_{\theta}(\mathcal{R}_n|Q_n). 
\end{equation} 
By optimizing $\mathcal{L}_{find}$, we maximize the probability of LLMs generating faithful plans $\mathcal{P}_n$ and the corresponding evidence paths $\mathcal{R}_n$ by distilling the knowledge from KGs.

\noindent\textbf{Answer predictor module optimization.} To optimize the answer predictor module, we aim to enable LLMs to conduct the final answer based on the evidence paths $\mathcal{R}_{n}$. By utilizing the evidence paths $\mathcal{R}_{N}$ formed from the $N$ sampled evidence paths to approximate the expectation, the objective function of reasoning optimization can be written as follows:

\begin{equation}
    \mathcal{L}_{reasoning} = log P_{\theta} (\mathcal{A}_n | Q_n, \mathcal{R}_{N}).
\end{equation}
The final objective function of EPERM is the combination of the finding and reasoning optimization, which can be formulated as:
\begin{equation}
\label{eq9}
    \mathcal{L} = \mathcal{L}_{find} + \mathcal{L}_{reasoning}.
\end{equation}
We use the same LLM for both the evidence path finder module and the answer predictor module, which are jointly trained on two instruction-tuning tasks. In this way, EPERM can better generate more accurate evidence reasoning paths and derive the final answers based on these evidence paths and their importance scores.

\section{Experiment}
In this section, we first introduce the experiment settings including datasets, baselines, and evaluation protocols. Secondly, we compare EPERM with competitive models and demonstrate its superiority. Thirdly, we conduct a series of ablation studies to analyze the importance of the three modules in the EPERM. Then, we analyze the impact of two important parameters on the proposed model. Finally, we do the case study to exploit how the EPERM finds the evidence paths and reasons the answers based on them.

\subsection{Experiment Setup}
\textbf{Datasets.} We evaluate the proposed EPERM on two benchmarks, WebQuestionSP (WebQSP)~\cite{yih2016value} and Complex WebQuestion (CWQ)~\cite{talmor2018web}, which contain up to 4-hop questions. The statistics of the datasets are given in Table~\ref{tab1}. Freebase~\cite{bollacker2008freebase} is the background knowledge graph for both datasets, which contains around 88 million entities, 20 thousand relations, and 126 million triples.

\begin{table}[htbp]
\caption{Statistics of datasets.}
\begin{tabular}{c|ccc}
\toprule
Datasets & \#Train & \#Test & Max \#hop \\ \midrule
WebQSP   & 2826    & 1628   & 2         \\
CWQ      & 27639   & 3531   & 4         \\ \midrule
\end{tabular}
\label{tab1}
\end{table}

\noindent \textbf{Evaluation Metrics.}
Following previous works~\cite{luo2023reasoning}, we use Hits@1 and F1 as the evaluation metrics. Hits@1 measures the proportion of questions whose top-1 predicted answer is correct. Since a question may correspond to multiple answers, F1 considers the coverage of all answers, which balances the precision and recall of the predicted answers.\\
\textbf{Baseline Models.}
We compare EPERM with the three types of KGQA methods. \textbf{Embedding based methods:} KV-Mem~\cite{miller2016key}, EmbedKGQA~\cite{saxena2020improving}, NSM~\cite{he2021improving}, TransferNet~\cite{shi2021transfernet}, KGT5~\cite{saxena2022sequence} and BAMnet~\cite{chen2019bidirectional}. \textbf{Retrieval based methods:} GraftNet~\cite{sun2018open}, GrailQA Ranking~\cite{gu2021beyond} PullNet~\cite{sun2019pullnet}, SR+NSM~\cite{zhang2022subgraph}, SR+NSM+E2E~\cite{zhang2022subgraph}, BeamQA~\cite{atif2023beamqa}. \textbf{LLM based methods:} LLaMA2-Chat-7B~\cite{touvron2023llama}, ChatGPT+CoT~\cite{luo2023reasoning}, EPR+NSM~\cite{ding2024enhancing}, UniKGQA~\cite{jiang2022unikgqa}, KD-CoT~\cite{wang2023knowledge}, RoG~\cite{luo2023reasoning}, StructGPT~\cite{jiang2023structgpt}, ToG+ChatGPT~\cite{sun2023think}
\\ \textbf{Implementations details.}
For EPERM, we use LLaMA2-Chat-7B~\cite{touvron2023llama} as the LLM backbone, which is instruction finetuned on the training split of WebQSP and CWQ as well as Freebase for 5 epochs. We generate the top-6 and top-5 weighted plans using beam-search for each question in WebQSP and CWQ respectively. Then it scores and filters out weighted evidence plans. For LLMs, we use zero-shot prompting to conduct KGQA. Our model is trained on 8 Nvidia A40 GPUs.
 
\begin{table*}[htbp] 
\caption{Performance comparison with different baselines on the two KGQA datasets. The best results are in bold and the second best results are underlined.
}
\centering
\begin{tabular}{c|ccccc}
\toprule
\multirow{2}{*}{\textbf{Type}}                   & \multirow{2}{*}{\textbf{Methods}} & \multicolumn{2}{c}{\textbf{WebQSP}} & \multicolumn{2}{c}{\textbf{CWQ}} \\ \cmidrule{3-6}
                                                 &                                   & \textbf{Hits@1} $\uparrow$   & \textbf{F1} $\uparrow$   & \textbf{Hits@1} $\uparrow$  & \textbf{F1} $\uparrow$ \\ \midrule
\multirow{6}{*}{\textbf{Embedding Based}}        & KV-Mem~\cite{miller2016key}                            & 46.7               & 34.5           & 18.4              & 15.7         \\
                                                 & EmbedKGQA~\cite{saxena2020improving}                         & 66.6               & -              & 45.9              & -            \\
                                                 & NSM~\cite{he2021improving}                               & 68.7               & 62.8           & 47.6              & 42.4         \\
                                                 & TransferNet~\cite{shi2021transfernet}                       & 71.4               & -              & 48.6              & -            \\
                                                 & KGT5~\cite{saxena2022sequence}                              & 56.1               & -              & 36.5              & -            \\
                                                  & BAMnet~\cite{chen2019bidirectional}                              & 55.6               & 51.8              & -              & -            \\\midrule
\multirow{6}{*}{\textbf{Retrieval Based}}        & GraftNet~\cite{sun2018open}                          & 66.4               & 60.4           & 36.8              & 32.7         \\              & GrailQA Ranking~\cite{gu2021beyond}                          & -               & 70.0           & -              & -         \\              & BeamQA~\cite{atif2023beamqa}                          & 73.3               & -           & -             & -         \\
                                                 & PullNet~\cite{sun2019pullnet}                           & 68.1               & -              & 45.9              & -            \\
                                                 & SR+NSM~\cite{zhang2022subgraph}                            & 68.9               & 64.1           & 50.2              & 47.1         \\
                                                 & SR+NSM+E2E~\cite{zhang2022subgraph}                        & 69.5               & 64.1           & 49.3              & 46.3         \\
                                                 & EPR+NSM~\cite{ding2024enhancing}                        & 71.2               & -           & 60.6               & -         \\\midrule

\multirow{9}{*}{\textbf{LLM Based}}       & LLaMA2-Chat-7B~\cite{touvron2023llama}                    & 64.4               & -              & 34.6              & -            \\
                                                 
                                                 & UniKGQA~\cite{jiang2022unikgqa}                           & 77.2               & \underline{72.2}           & 51.2              & 49.1         \\
                                                 & KD-CoT~\cite{wang2023knowledge}                            & 68.6               & 52.5           & 55.7              & -            \\
                                                 & ChatGPT+CoT~\cite{luo2023reasoning}                               & 75.6               & -           & 48.9              & -         \\
                                                 & RoG~\cite{luo2023reasoning}                               & \underline{85.7}               & 70.8           & \underline{62.6}              & \underline{56.2}        \\
                                                 & StructGPT~\cite{jiang2023structgpt}                         & 72.6               & –              & –                 & –            \\
                                                 & ToG+ChatGPT~\cite{sun2023think}                       & 76.2               & –              & 58.9              & –            \\  \cmidrule{2-6} 
                                                 & EPERM (Ours)                              & \textbf{88.8}                   &  \textbf{72.4}               & \textbf{66.2}                   & \textbf{58.9}              \\ 
\bottomrule
\end{tabular}
\label{tab2}
\end{table*}

\subsection{Main Results}
We present the main results on two KGQA datasets(CWQ, WebQSP) in Table~\ref{tab2}. Our observations based on the results are as follows. First, retrieval-based approaches outperform embedding-based methods by retrieving relevant subgraphs from KGs, which reduces reasoning complexity. Furthermore, SR+NSM and SR+NSM+E2E adopt relation paths-based retrieval which achieves better performance, highlighting the importance of relation paths. Compared to these two types of traditional methods, EPERM achieves remarkable improvement across all metrics on two datasets. Specifically, it achieves a 19.3\% (27.7\% relative), and 16.0\% (31.8\% relative) increase in Hits@1 on the WebQSP and CWQ respectively. Second, compared to the methods of jointly using KGs and LLMs, EPERM still achieves improvement across all metrics on two datasets. Specifically, it achieves a 3.1\% (3.6\% relative), and 3.6\% (5.8\% relative) increase in Hits@1 scores over the SOTA model on the WebQSP and CWQ respectively. In conclusion, these results illustrate that by decomposing the KGQA task into three stages, EPERM is able to find the more accurate evidence paths that are highly relevant to the questions and have different weights to help the reasoning stage, assisting the answer predictor to achieve better reasoning performance.

\subsection{Ablation Study}
First, we conduct a series of ablation studies to analyze the importance of the weighted evidence paths for the performance of the subsequent answer predictor. We compare three variants: 1) w/o evidence path finder, where we remove the evidence path finder and perform the answer predictor directly. 2) w/o scoring and filtering out the evidence paths in the evidence path finder, where we do not filter the evidence path by weighted plans. The results are shown in Table~\ref{tab3}. Based on the results, it is obvious that the performance of the answer predictor will be greatly reduced if we remove the evidence path finder. This is because the input is solely the question, causing the model to degrade into LLM that directly answers the questions. Additionally, if we do not score and filter out the evidence paths during the evidence path finding stage, it will also lead to a decrease in the final performance of the answer predictor. Further scoring and filtering of evidence paths can take into account the varying contributions of different evidence paths to question reasoning. All these results demonstrate the effectiveness of weighted evidence paths for the performance of the subsequent answer predictor.

\noindent Second, to analyze the importance of the answer predictor, we remove the answer predictor and use all answers from the weighted evidence paths as results. The results are shown in Table~\ref{tab4}. From these results, it can be inferred that the answer predictor can further infer and judge from the weighted evidence paths, and obtain more accurate results. Although removing the answer predictor leads to a high recall rate due to an increased number of answers, precision significantly drops. 

\begin{figure}[htbp]
\caption{The Hit@1 scores of EPERM with the total number of generated plans $s$ and the number of Top-$S$ path filtered in every hop.}
\subfloat[Comparison on WebQSP]
{\includegraphics[height=0.36\textwidth]{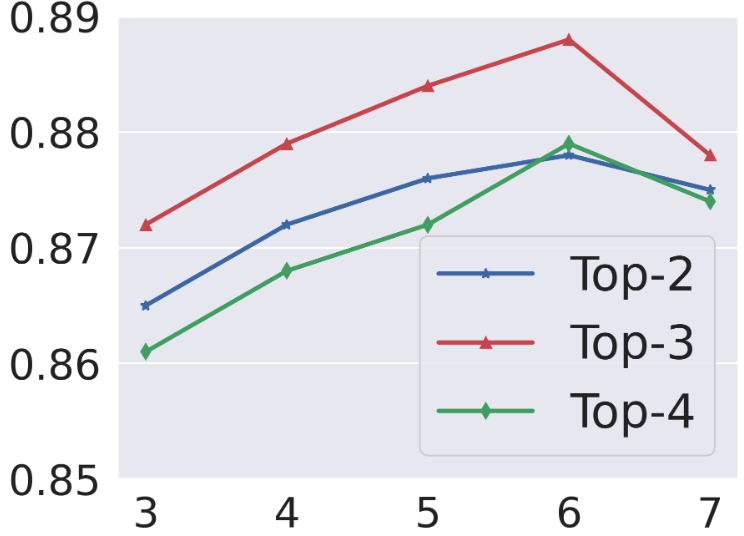}}
\subfloat[Comparison on CWQ]
{\includegraphics[height=0.36\textwidth]{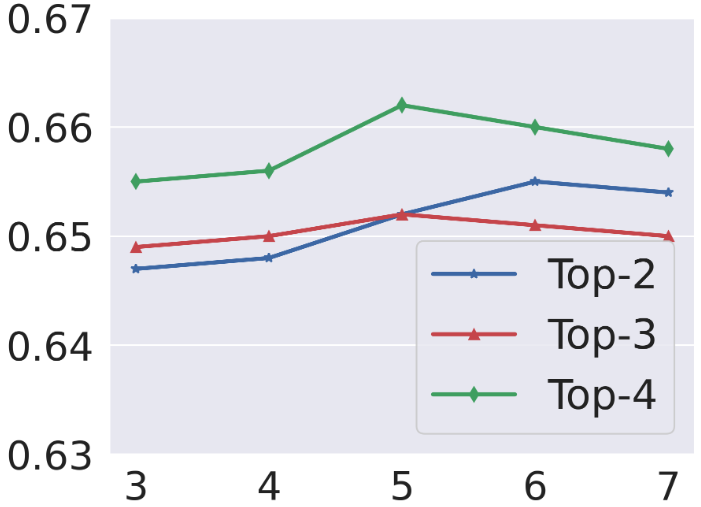}}
\label{fig3}
\end{figure}

\begin{table}[htbp]
\caption{Ablation on the evidence path finder module.}
\begin{tabular}{c|cccc}
\toprule
\multirow{2}{*}{\textbf{Methods}}                                         & \multicolumn{2}{c}{\textbf{WebQSP}} & \multicolumn{2}{c}{\textbf{CWQ}} \\ \cmidrule{2-5} 
                                                                          & \textbf{Hits@1}    & \textbf{F1}    & \textbf{Hits@1}   & \textbf{F1}  \\ \midrule
EPERM                                                                   & 88.8               & 72.4           & 66.2              & 58.9         \\ \midrule
\begin{tabular}[c]{@{}c@{}}EPERM w/o \\ filtering path  \end{tabular}        & 84.2               & 69.1              & 61.3              & 55.8            \\ \midrule
\begin{tabular}[c]{@{}c@{}}EPERM w/o \\ evidence path \end{tabular} & 66.2               &50.3                & 36.8                  & 35.7             \\ \bottomrule
\end{tabular}
\label{tab3} 
\end{table}

\begin{table}[htbp]
\caption{Ablation on the answer predictor module.}
\begin{tabular}{c|cccc}
\toprule
\multirow{2}{*}{\textbf{Methods}}                                         & \multicolumn{2}{c}{\textbf{WebQSP}} & \multicolumn{2}{c}{\textbf{CWQ}} \\ \cmidrule{2-5} 
                                                                          & \textbf{Hits@1}    & \textbf{Recall}    & \textbf{Hits@1}   & \textbf{Recall}  \\ \midrule
EPERM                                                                   & 88.8               & 76.4           & 66.2              & 60.9         \\ \midrule
\begin{tabular}[c]{@{}c@{}}EPERM w/o \\ answer predictor\end{tabular}        &62.3                &79.8              &33.1               &66.2             \\ \bottomrule
\end{tabular}
\label{tab4}
\end{table}

\subsection{Influence of hyperparameters}
In this subsection, we conduct two experiments to analyze the impact of the total number of generated plans $s$ and the number of filtering paths $S$ on the proposed model. Firstly, we change the number of generated plans $s$. From the figure~\ref{fig3}, it can be inferred that when we fix the number of filtering paths $S$, as the number of plans increases, the performance of the model initially rises and then falls. This is because when $s$ is too small, the coverage rate of the answers is low, making it difficult to cover all correct answers. As $s$ increases, more relevant information to the query is retrieved, leading to a higher answer coverage rate and improved model performance. However, as $s$ continues to increase, it introduces unnecessary noise, which can degrade the performance of the model. In addition, an appropriate $s$ is of great significance to the model's performance. Specifically, for the WebQSP, the best $s$ is 6. While for the CWQ the best $s$ is 5. Secondly, we change the number of filtering top $S$ paths in every hop. We can see that an appropriate filtering path number $S$ plays a crucial role in the model's performance. A smaller $S$ potentially removes more irrelevant information but also risks discarding correct information along the way. Conversely, a larger $S$ might introduce noise and degrade the performance of the model. Typically, in the WebQSP dataset, the suitable $S$ is 3, whereas a value of 4 is appropriate for the CWQ dataset.

\begin{figure}[htbp]
\centering 
{\includegraphics[height=0.675\textwidth]{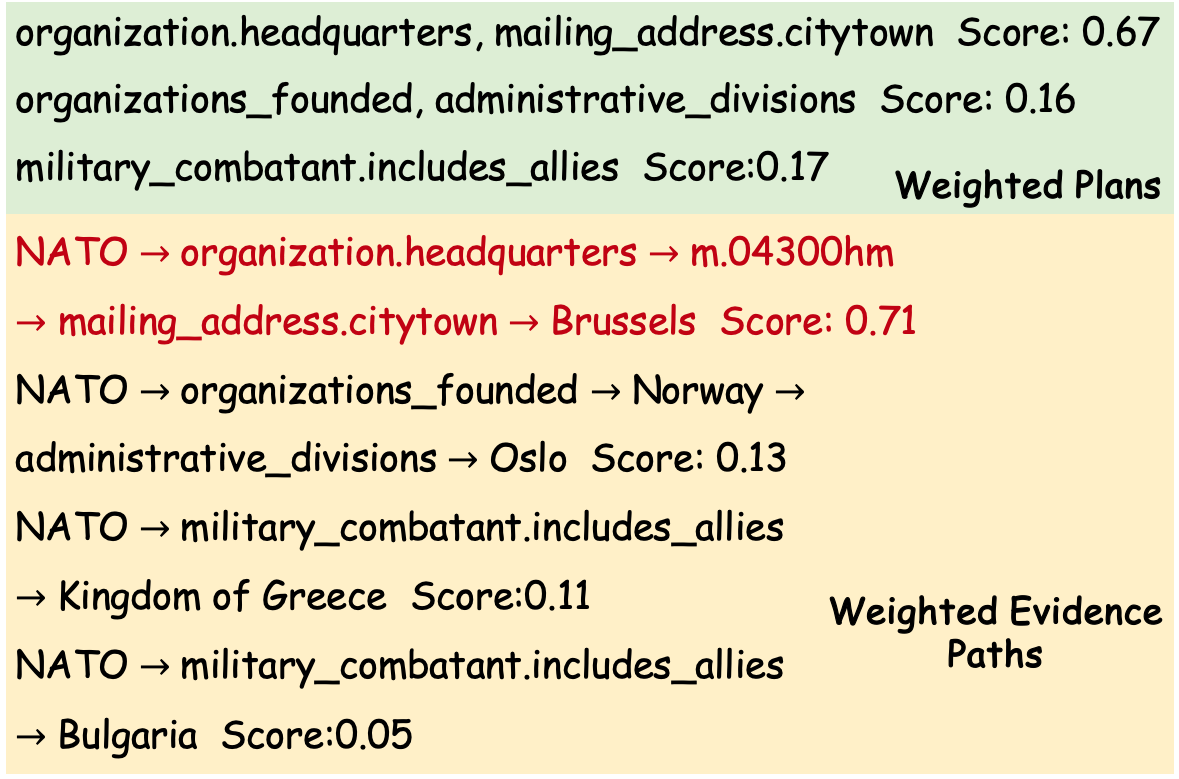}}
\caption{Example of EPERM reasoning based on weighted evidence paths.} 
\label{fig4}
\end{figure}

\subsection{Case Study}
Finally, we explore how the EPERM reasoning the answers based on the weighted evidence paths. We illustrate a case study in Figure~\ref{fig4}. We can see that for the question "Where are the NATO headquarters located?", EPERM can generate a series of weighted plans and then it scores and filters out the weighted evidence paths in the subgraph based on the weighted plans. Although these paths are related to the problem, they still have different confidence scores to reason the questions. If we treat each path equally, it will degrade the reasoning performance. For example, the first path ``${\rm NATO}\xrightarrow{\rm organization.headquarters}\rm m.04300hm \xrightarrow{\rm mailing\_address.citytown} \rm Brussels$'' is more likely to reason the final result. Because the question emphasizes where NATO's headquarters is located. The other evidence paths e.g., ``${\rm NATO}\xrightarrow{\rm organizations\_founded}\rm Norway \xrightarrow{\rm administrative\_divisions} \rm Oslo$'' focus on where NATO's various departments are located, which are supposed to have lower confidence in reasoning the answer. In this way, the Answer Predictor in the EPERM can better make the final choice in the reasoning stage.

\section{Conclusion}
In this paper, we propose a novel framework called the Evidence Path Enhanced Reasoning Model (EPERM) to address RAG-based knowledge graph question answering tasks. This framework explores the integration of the generative and reasoning capabilities of large language models (LLMs) with prior knowledge in knowledge graphs for faithful reasoning. We reformulate the KGQA task as a graphical model comprising three stages. In the first stage, EPERM utilizes a fine-tuned LLM to retrieve a subgraph related to the question from the original knowledge graph. In the second stage, the evidence path finder generates a series of weighted plans that reliably support the reasoning process. It then scores and filters the weighted evidence paths within the subgraph based on these plans. Finally, in the third stage, the answer predictor leverages the weighted evidence path to reason the final answer. Since the weight of each evidence path indicates the different importance of the structural information for reasoning the question, EPERM can better leverage them to reason the answer. Extensive experiments on benchmark datasets demonstrate that EPERM achieves superior performances in KGQA tasks.

\section*{Acknowledgements}
This work was supported by the National Natural Science Foundation of China under Grant No.U20B2070, No.61976199.

\bibliography{aaai25}

\clearpage
\appendix
\section*{Appendix}

\subsection{A.1 Details of EPERM}
We introduce the details of directed probabilistic graphical model of KGQA tasks in this section.

\begin{figure}[htbp]
\centering
{\includegraphics[height=0.4\textwidth]{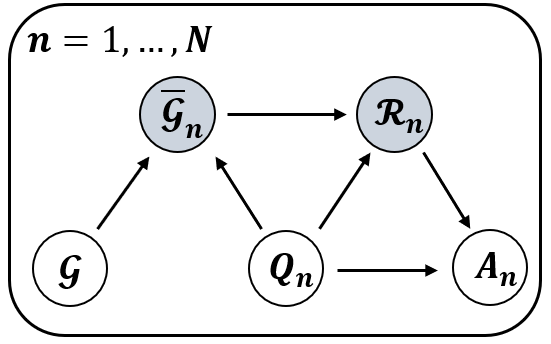}}
\caption{The directed graphical model of KGQA tasks.}
\label{p1}
\end{figure}

\noindent Given a question $Q_n$ and its answers $\mathcal{A}_n$, we formalize the KBQA problem as to model the probabilistic distribution $P(\mathcal{A}_n|\mathcal{G}, Q_n)$. We introduce two latent variables: an evident subgraph $\overline{\mathcal{G}}$ and a series of evident paths $\mathcal{R}_n$ related to the question $Q_n$. Then the KGQA task can be reformulated as a probabilistic graphical model in Figure~\ref{p1}. We follow the d-separation principle. Since the $\mathcal{R}_n\perp\!\!\!\perp\mathcal{G} | \overline{\mathcal{G}}$, $\mathcal{A}_n\perp\!\!\!\perp\mathcal{G} | \mathcal{R}_n$ and $\mathcal{A}_n\perp\!\!\!\perp\overline{\mathcal{G}} | \mathcal{R}_n$. According to the graph, the objective distribution $P(\mathcal{A}_n|\mathcal{G}, Q_n)$ can be reformulated as below:

\begin{align}
\label{eq1}
    & P_{\theta}(\mathcal{A}_n|\mathcal{G}, Q_n) \nonumber \\ & = \Sigma_{\overline{\mathcal{G}}}P_{\theta}(\mathcal{A}_n, \overline{\mathcal{G}}|\mathcal{G}, Q_n) \nonumber \\
    & = \Sigma_{\overline{\mathcal{G}}} P_{\theta}(\mathcal{A}_n|\mathcal{G}, \overline{\mathcal{G}}, Q_n) P_{\theta}(\overline{\mathcal{G}}|\mathcal{G}, Q_n) \nonumber \\ 
    & = \Sigma_{\mathcal{R}_n}\Sigma_{\overline{\mathcal{G}}}P_{\theta}(\mathcal{A}_n, \mathcal{R}_n|\mathcal{G}, \overline{\mathcal{G}}, Q_n)P_{\theta}(\overline{\mathcal{G}}|\mathcal{G}, Q_n) \nonumber \\ &=\Sigma_{\mathcal{R}_n}\Sigma_{\overline{\mathcal{G}}}P_{\theta}(\mathcal{A}_n |\mathcal{R}_n, \mathcal{G}, \overline{\mathcal{G}}, Q_n) \nonumber \\ 
    & \quad  P_{\theta}(\mathcal{R}_n | \mathcal{G}, \overline{\mathcal{G}}, Q_n)  P_{\theta}(\overline{\mathcal{G}}|\mathcal{G}, Q_n) \nonumber \\
    &= \Sigma_{\mathcal{R}_n} \Sigma_{\overline{\mathcal{G}}} P_{\theta}(\mathcal{A}_n|\mathcal{R}_n, Q_n)  P_{\theta}(\mathcal{R}_n|\overline{\mathcal{G}}, Q_n) P_{\theta}(\overline{\mathcal{G}}|\mathcal{G}, Q_n). \nonumber
\end{align} 
This is the equation 1 in the main text. In the above equation, the proposed EPERM can be divided into three parts.

\subsection*{A.2 \ \ Details of Optimization}
Next, we introduce the details of the evidence lower bound (ELBO) in the main text. The objective function in equation 1
can be optimized by maximizing the equation:

\begin{align}
    & log P_{\theta}(\mathcal{A}_n|\mathcal{G}, Q_n) \nonumber \\ 
    & = log P_{\theta}\Sigma_{\mathcal{R}_n} \Sigma_{\overline{\mathcal{G}}} P_{\theta}(\mathcal{A}_n|\mathcal{R}_n, Q_n)  P_{\theta}(\mathcal{R}_n|\overline{\mathcal{G}}, Q_n) P_{\theta}(\overline{\mathcal{G}}|\mathcal{G}, Q_n) \nonumber \\
    & = log \Sigma_{\mathcal{R}_n} \Sigma_{\overline{\mathcal{G}}} P_{\theta}(\mathcal{A}_n|\mathcal{R}_n, Q_n)  P_{\theta}(\mathcal{R}_n|\overline{\mathcal{G}}, Q_n) P_{\theta}(\overline{\mathcal{G}}|\mathcal{G}, Q_n) \nonumber \\
    & \quad \frac{q_{1}(\mathcal{R}_n | \mathcal{A}_n) q_{2}(\overline{\mathcal{G}}|\mathcal{G}, Q_n)}{q_{1}(\mathcal{R}_n | \mathcal{A}_n) q_{2}(\overline{\mathcal{G}}|\mathcal{G}, Q_n)} \nonumber \\
    & = log \mathbb{E}_{\mathcal{R}_n \sim q_1}  \frac{P_{\theta}(\mathcal{R}_n | \mathcal{A}_n)}{q_{1}(\mathcal{R}_n | \mathcal{A}_n)}\mathbb{E}_{\overline{\mathcal{G}} \sim q_2} \frac{P_{\theta}(\mathcal{R}_n|\overline{\mathcal{G}}, Q_n) P_{\theta}(\overline{\mathcal{G}}|\mathcal{G}, Q_n)}{q_{2}(\overline{\mathcal{G}}|\mathcal{G}, Q_n)} \nonumber \\
    & \ge \mathbb{E}_{\mathcal{R}_n \sim q_1} log (\frac{P_{\theta}(\mathcal{R}_n | \mathcal{A}_n)}{q_{1}(\mathcal{R}_n | \mathcal{A}_n)} \mathbb{E}_{\overline{\mathcal{G}} \sim q_2}\frac{P_{\theta}(\mathcal{R}_n|\overline{\mathcal{G}}, Q_n) P_{\theta}(\overline{\mathcal{G}}|\mathcal{G}, Q_n)}{q_{2}(\overline{\mathcal{G}}|\mathcal{G}, Q_n)}) \nonumber \\
    & = \mathbb{E}_{\mathcal{R}_n \sim q_1} log \frac{P_{\theta}(\mathcal{R}_n | \mathcal{A}_n)}{q_{1}(\mathcal{R}_n | \mathcal{A}_n)} + \nonumber \\ 
    & \quad \mathbb{E}_{\mathcal{R}_n \sim q_1}log(\mathbb{E}_{\overline{\mathcal{G}} \sim q_2} \frac{P_{\theta}(\mathcal{R}_n|\overline{\mathcal{G}}, Q_n) P_{\theta}(\overline{\mathcal{G}}|\mathcal{G}, Q_n)}{q_{2}(\overline{\mathcal{G}}|\mathcal{G}, Q_n)})  \nonumber \\
    & \ge \mathbb{E}_{\mathcal{R}_n \sim q_1} log (\frac{P_{\theta}(\mathcal{R}_n | \mathcal{A}_n)}{q_{1}(\mathcal{R}_n | \mathcal{A}_n)} + \nonumber \\
    & \quad \mathbb{E}_{\mathcal{R}_n \sim q_1} \mathbb{E}_{\overline{\mathcal{G}} \sim q_2} log (\frac{P_{\theta}(\mathcal{R}_n|\overline{\mathcal{G}}, Q_n) P_{\theta}(\overline{\mathcal{G}}|\mathcal{G}, Q_n)}{q_{2}(\overline{\mathcal{G}}|\mathcal{G}, Q_n)}) \nonumber \\
    & = \mathbb{E}_{\mathcal{R}_n \sim q_1}logP(\mathcal{A}_n | \mathcal{R}_n)+ \nonumber \\
    &  \quad  \mathbb{E}_{\mathcal{R}_n \sim q_1} [\mathbb{E}_{\overline{\mathcal{G}} \sim q_2} logP_{\theta}(\mathcal{R}_n|\overline{\mathcal{G}}, Q_n) - logq_{1}(\mathcal{R}_n | \mathcal{A}_n)] -   \nonumber \\
    & \quad \mathbb{E}_{\mathcal{R}_n \sim q_1} \mathbb{E}_{\overline{\mathcal{G}} \sim q_2} log(\frac{q_{2}(\overline{\mathcal{G}}|\mathcal{G}, Q_n)}{P_{\theta}(\overline{\mathcal{G}}|\mathcal{G}, Q_n)}) \nonumber \\
    & = \mathbb{E}_{\mathcal{R}_n \sim q_1}[logP(\mathcal{A}_n | \mathcal{R}_n)- D_{KL}(q_{2}(\overline{\mathcal{G}}|\mathcal{G}, Q_n) || P(\overline{\mathcal{G}}|\mathcal{G}, Q_n))] \nonumber \\ 
    & \quad -\mathbb{E}_{\overline{\mathcal{G}} \sim q_2}[D_{KL}(q_{1}(\mathcal{R}_n|A_n) || P(\mathcal{R}_n|\overline{\mathcal{G}}, Q_n))] \nonumber 
\end{align}

\subsection*{A.3 \ \ Statistics of Datasets}
We adopt two benchmark KGQA datasets: WebQuestionSP (WebQSP) and Complex WebQuestions (CWQ) in this work. We follow previous works to use the same train and test splits for fair comparison. The statistics of the answer numbers and reasoning hops are presented in Table 1 and Table 2.

\begin{table}[h]
\caption{Statistics of the number of answers for questions in two datasets.}
\begin{tabular}{c|ccc}
\toprule
Datasets & \#Ans $=$ 1 & 2 $\ge$ \#Ans $\le$ 4 & \#Ans $\ge$ 5 \\ \midrule
WebQSP   &  51.2\%    & 27.4\%   & 20.4\%         \\
CWQ      &  70.6\%    & 19.4\%   & 10.0\%         \\ \midrule
\end{tabular}
\label{tab1}
\end{table}

\begin{table}[h]
\caption{Statistics of the hop of questions in two datasets.}
\begin{tabular}{c|ccc}
\toprule
Datasets &1 hop  & 2 hop  & 3 hop  \\ \midrule
WebQSP   &   65.49\%    & 34.51\%   & 0.00\%             \\
CWQ      &  40.91\%    & 38.34\%   &  20.75\%            \\ \midrule
\end{tabular}
\label{tab2}
\end{table}

\subsection*{A.4 \ \ Details of prompts in three modules}
\begin{figure*}[ht]
\begin{center}
\begin{tcolorbox}[colback=gray!10, colframe=black!70, title=Prompt for scoring the relations in subgraph retriever module]

Instruction:

Assume you are a **semantic analysis expert**. You will receive an encyclopedic question, related topic entities, and a set of several retrieved relationships that need to be filtered (which need to assist in inferring the question). Your task is to carefully consider the information needed to reason about the question and, based on the semantics of the existing reasoning path, select the top {{budget}} relationships from the set that are most likely to help infer the answer to the question. \\

Guidelines

1. The format of the input is: \\
   **Question**: \\
   The input question \\
   **Topic entity**:  \\
   The related topic entities \\
   **Several retrieved relationships**: \\ 
   A set of several retrieved relationships separated by semicolons (;) \\
   **Budget**: \\
   The number of the selected relationships that are most likely to infer the result.
   
2. The number of the selected relationships are no more than $\{\{budget\}\}$. reset counter between $<count>$ and $</count>$ to $\{\{budget\}\}$.

3. You are allowed to select $\{\{budget\}\}$ relationships (starting budget), keep track of it by counting down within tags $<count>$ $</count>$, STOP GENERATING MORE RELATIONSHIPS when hitting 0.

4. Please provide your count, reasons, scores, and selected relationships in the following XML format.

   $<count>$ [starting budget] $</count>$
   
   $<choice>$ The relationship you select that is most likely to infer the question. $</choice>$
   
   $<reason>$ Provide the reasons for the score you assigned to the relationship 1 for helping infer the questions. $</reason>$
   
   $<score>$ The confidence score 0.0-1.0 to select this relation $</score>$
   
   $<count>$ [remaining budget] $</count>$
   
   $<choice>$ The 2-th relationship you select that is likely to infer the questions. $</choice>$
   
   $<reason>$ Provide the reasons for the score you assigned to the relationship 2 for helping infer the questions. $</reason>$
   
   $<score>$ The confidence score 0.0-1.0 to select this relationship $</score>$
   
   ...
   
   $<count>$ 1 $</count>$
   
   $<choice>$ The $\{\{budget\}\}$-th relationship you select that is likely to infer the questions. $</choice>$
   
   $<reason>$ Provide the reasons for the score you assigned to the relationship $\{\{budget\}\}$ for helping infer the questions. $</reason>$
   
   $<score>$ The confidence score 0.0-1.0 to select this relationship $</score>$\\

Input:

**Question**:

\{question\}

**Topic entity**:

\{topic\_entity\}

**Several retrieved relationships**:

\{relation\}

**Budget**:

\{budget\} \\

Output:
\end{tcolorbox}
\end{center}
\end{figure*}

\begin{figure*}[ht]
\begin{center}
\begin{tcolorbox}[colback=gray!10, colframe=black!70, title=Prompt for scoring entity candidates in path finder module]

Instruction:

I want to answer the question through a relationship path. There will be multiple candidate entities along the path. Please help me choose the entity that can better infer the answer to the question.

\#EXAMPLE\#

Question: 

who did cristiano ronaldo play for in 2010?

Plans: 

soccer.football\_player\_match\_participation.player, soccer.football\_player\_match\_participation.team

Candidate Entity: 

m.0g9lr08

Relevent Information: 

m.0g9lr08 $\to$ soccer.football\_player\_match\_participation.match $\to$ 2010 FIFA World Cup Group G - POR ./. PRK

m.0g9lr08 $\to$ soccer.football\_player\_match\_participation.team $\to$ Portugal national football team

m.0g9lr08 $\to$ soccer.football\_player\_match\_participation.shirt\_number $\to$ 7

m.0g9lr08 $\to$ soccer.football\_player\_match\_participation.part\_of\_starting\_lineup $\to$ true 

Inference criteria: 

To score the contribution of entities to the question, we need to determine if the entities are relevant to the question based on provided relevant information. For example, 
if the relevant information of entity m.0g9lr08 includes the 2010 FIFA World Cup, which is consistent with the time information in the question, there is a higher likelihood 
that this entity is relevant to the question. If the relevant information of an entity is not consistent with the time information "2010" in the question, the relevance between 
that entity and the question is lower, and the score will be lower such as 0.56. Based on the above speculative evidence, the scores of m.0g9lr08 could be assigned 0.93(high relevance).

INPUT

Question:

\{\}

Plans:

\{\}

Candidate Entity:

\{\}

Relevent Information:

\{\}

Please help me score the relevance between the Candidate Entity \{\} and the question based on the relevent information of the Candidate Entity in the INPUT. The scoring range is from 0 to 1 and different entity has different socre. Finally, please output the different socre in the following format: \{\{Candidate Entity:\{\},Score:xxx\}\}, and do not output explanations.

\end{tcolorbox}
\end{center}
\end{figure*}

\begin{figure*}[ht]
\begin{center}
\begin{tcolorbox}[colback=gray!10, colframe=black!70, title=Prompt in answer predictor module]

Instruction:

Please answer the question based on the scores of the reasoning paths and return all possible answers in a list. Each path has a score (0.0-1.0) at its end, and please strictly adhere to the prediction criterion of outputting the entity with a high score.

\#\#EXAMPLE

Reasoning Paths and Scores:

C $\to$ tv.tv\_character.appeared\_in\_tv\_episodes $\to$ m.09p1747 $\to$ tv.tv\_actor.guest\_roles $\to$ A Score:0.102

C $\to$ tv.tv\_character.appeared\_in\_tv\_episodes $\to$ m.0jzvxtw $\to$ tv.tv\_actor.guest\_roles $\to$ A Score:0.22

C $\to$ tv.tv\_character.appeared\_in\_tv\_episodes $\to$ g.11byb39pmc $\to$ tv.tv\_actor.guest\_roles $\to$ A Score:0.122

C $\to$ tv.regular\_tv\_appearance.character $\to$ m.04d4q86 $\to$ tv.tv\_actor.starring\_roles $\to$ B Score:0.322

C $\to$ tv.regular\_tv\_appearance.character $\to$ m.0k6pxpv $\to$ tv.tv\_actor.starring\_roles $\to$ B Score:0.312

Question: 
who was the original voice of C in the movie?

Output: 
The answer is B

Reasoning Paths and their scores:

\{\}

Question:

\{\}
\end{tcolorbox}
\end{center}
\end{figure*}

\end{document}